\pdfoutput=1

\documentclass[11pt, preprint]{article}

\usepackage[preprint]{acl}

\usepackage{times}
\usepackage{latexsym}
\usepackage{pifont}
\usepackage{enumitem}
\usepackage{multirow}
\usepackage{booktabs}
\usepackage{float}

\usepackage[T1]{fontenc}

\usepackage[utf8]{inputenc}

\usepackage{microtype}

\usepackage{inconsolata}

\usepackage{graphicx}
\usepackage{amsmath}

\title{Where You Go is Who You Are: Behavioral Theory-Guided LLMs for Inverse Reinforcement Learning}

\author{
 \textbf{Yuran Sun\textsuperscript{1}},
 \textbf{Susu Xu\textsuperscript{2}},
 \textbf{Chenguang Wang\textsuperscript{2,3}},
 \textbf{Xilei Zhao\textsuperscript{1}\thanks{Corresponding author: \texttt{xilei.zhao@essie.ufl.edu}}}
\\
 \textsuperscript{1}University of Florida,
 \textsuperscript{2}Johns Hopkins University,
 \textsuperscript{3}Stony Brook University
}

\begin{document}
\maketitle

\begin{abstract}
Big trajectory data hold great promise for human mobility analysis, but their utility is often constrained by the absence of critical traveler attributes, particularly sociodemographic information. While prior studies have explored predicting such attributes from mobility patterns, they often overlooked underlying cognitive mechanisms and exhibited low predictive accuracy. This study introduces \textit{SILIC}, short for Sociodemographic Inference with LLM-guided Inverse Reinforcement Learning (IRL) and Cognitive Chain Reasoning (CCR), a theoretically grounded framework that leverages LLMs to infer sociodemographic attributes from observed mobility patterns by capturing latent behavioral intentions and reasoning through psychological constructs. Particularly, our approach explicitly follows the Theory of Planned Behavior (TPB), a foundational behavioral framework in transportation research, to model individuals' latent cognitive processes underlying travel decision-making. The LLMs further provide heuristic guidance to improve IRL reward function initialization and update by addressing its ill-posedness and optimization challenges arising from the vast and unstructured reward space. Evaluated in the 2017 Puget Sound Regional Council Household Travel Survey, our method substantially outperforms state-of-the-art baselines and shows great promise for enriching big trajectory data to support more behaviorally grounded applications in transportation planning and beyond.
\end{abstract}

\section{Introduction}
Understanding human mobility patterns is critical for many fields, such as transportation engineering \cite{luca2021survey}, marketing \cite{ghose2019mobile}, urban planning \cite{haraguchi2022human}, and emergency management \cite{yabe2016estimating, zhao2022estimating}. In recent years, researchers and practitioners have been leveraging real-world trajectory data generated by mobile devices (or synthetic trajectories \cite{zhu2023synmob, kim2024privacy}) to analyze and model people's movements to facilitate decision-making \cite{ghose2019mobile, zhao2022estimating}. However, these trajectory datasets often lack traveler attributes, particularly sociodemographic information, limiting their usefulness for important applications such as causal analysis to inform transportation planning and policy \cite{holz2019land}, personalized incentives for targeted marketing \cite{zhong2015you}, and tailored crisis communication to enhance evacuation compliance \cite{leykin2016leveraging}.

Some prior studies have explored inferring sociodemographic attributes from people's travel trajectories, e.g., \citet{chen2024coupling, zhang2024social}. These studies typically used classical machine learning models (e.g., Support Vector Machines, XGBoost) to predict attributes like gender, age, or income from features extracted from individual trajectories \cite{zhu2017prediction, wu2019inferring, bakhtiari2023inferring}. While insightful, these approaches often exhibit limited predictive performance, reducing their effectiveness in large-scale real-world applications.

\begin{figure*}[t]
    \centering
    \includegraphics[width=0.8
    \textwidth]{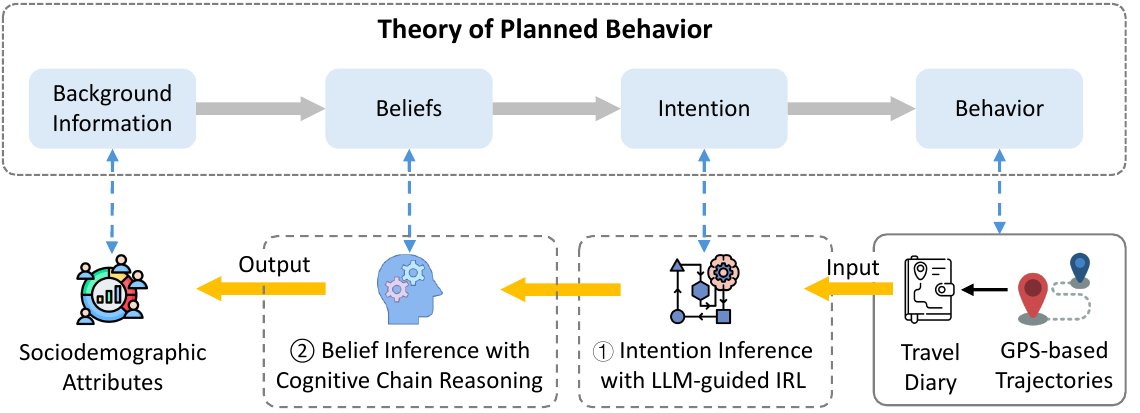}
    \vspace{-0.5em}
    \caption{Overview of our proposed framework. We inversely follow the Theory of Planned Behavior (TPB) to predict sociodemographic attributes from travel trajectories. Our method first uses an LLM-guided IRL model to infer behavioral intentions, followed by a Cognitive Chain Reasoning strategy that predicts sociodemographic attributes via intermediate belief constructs. }
    \label{fig:TPB}
    \vspace{-1em}
\end{figure*}

A key reason for their poor performance is the oversimplification of the link between sociodemographic attributes and mobility, often ignoring \textit{underlying cognitive processes}. Human mobility is complex and shaped by latent cognitive processes, often modeled using the Theory of Planned Behavior (TPB) \cite{ajzen1991theory}, as shown in Figure \ref{fig:TPB} (upper section). TPB posits that background factors, such as sociodemographic and contextual attributes, inform three core beliefs: attitudes, subjective norms, and perceived behavioral control \cite{ajzen2020theory}. These beliefs collectively shape individuals’ intentions, which ultimately drive their observable behaviors (e.g., travel decisions). Therefore, predicting sociodemographic attributes directly from mobility patterns, without accounting for these mediating constructs, may lead to inaccurate or incomplete inferences.

Addressing this limitation requires models capable of reasoning over latent cognitive constructs that mediate behavior—a capacity demonstrated by large language models (LLMs) \cite{nguyen2024predicting, lee2024reasoning, liu2025dataset}. Unlike traditional machine learning models, LLMs exhibit strong reasoning and theory-of-mind (ToM) abilities \cite{street2024llm, zhang2024llm, bandyopadhyay2025thinking}. However, predicting sociodemographic attributes from mobility patterns requires inversely following the TPB framework. Leveraging existing LLMs for this inverse cognitive modeling still face the following challenges: (1) \textbf{Misalignment with behavioral theory}: Without structural guidance \cite{zhou2023far}, LLMs often capture only partial reasoning patterns, leading to incomplete or theoretically inconsistent interpretations \cite{tjuatja2024llms}; (2) \textbf{Lack of iterative behavioral refinement}: Although LLMs can perform inverse alignment to infer latent mental states from behavior, they often generate intentions in a single pass without feedback refinement \cite{sun2improving, sun2024inverse}; and (3) \textbf{Ambiguity in mapping mental states to identity}: Similar mental states can arise across different sociodemographic groups due to shared external influences, making LLMs prone to misclassification without contextual input \cite{chen2025perceptions}.

Based on these challenges, we raise a key question: \textit{How can we construct a principled framework that efficiently leverages LLMs to inversely follow the cognitive pathway in the TPB to recover sociodemographic attributes?}

To address this research question, we propose \textit{SILIC}, a two-stage framework, as illustrated in Figure~\ref{fig:TPB} (bottom section). In the first stage, we propose an LLM-guided Inverse Reinforcement Learning (IRL) approach to infer individualized reward functions \cite{ng2000algorithms} that reflect intentions \cite{liang2025analyzing} from travel trajectories (Figure \ref{fig:TPB}, dashed box denoted as \ding{172}). Our core idea is to address the challenge of limited iterative behavioral refinement in LLMs by integrating IRL. In turn, we leverage structured domain knowledge from LLMs \cite{wu2024reward} to guide both the initialization and update of the IRL reward function \cite{ma2023eureka, kwon2023reward, chu2023accelerating}. Such guidance is essential for addressing key methodological limitations of IRL, including: (1) \textbf{Ill-posed nature} \cite{ng2000algorithms, cao2021identifiability}, where, in the context of travel behavior analysis, multiple latent intentions can explain the same observed travel patterns; (2) \textbf{Vast and unstructured reward space}, which hinders efficient exploration and increases the risk of converging to behaviorally implausible or non-generalizable solutions \cite{adams2022survey}; and (3) \textbf{Suboptimal reward convergence}, particularly during the early training stages when lacking heuristic guidance \cite{wu2024reward}.

In the second stage, we introduce a Cognitive Chain Reasoning (CCR) strategy to predict the final sociodemographic attributes from inferred intentions (Figure \ref{fig:TPB}, dashed box denoted as \ding{173}). Leveraging LLMs’ ToM capacities, This stage is designed to guide the LLM in aligning with the theoretical framework by first inferring belief constructs, and subsequently mapping them to sociodemographic attributes.. By incorporating relevant contextual factors (e.g., urbanicity, transit access) in the latter step, the model mitigates the ambiguity in LLM-based mappings and enables predictions that are both theory-aligned and context-aware\cite{tjuatja2024llms, petrov2024limited, chen2025perceptions}.

In summary, our major contributions are:
\begin{itemize}[leftmargin=*,topsep=-0.3em,itemsep=-0.3em]
\item We propose a novel methodological framework that inversely follows TPB to predict sociodemographic attributes from travel trajectories. 
\item We leverage LLMs to provide heuristic support for IRL reward function initialization and updates, enabling the inference of individualized and well-posed reward solutions.
\item We introduce a CCR strategy that guides the LLM to make predictions in full alignment with the TPB, by explicitly modeling the mediating cognitive constructs.
\item The experiments demonstrate that, compared to baselines, our model achieves substantially higher accuracy (e.g., a 30.93\% improvement in gender prediction). By inferring sociodemographic attributes from trajectories, we can enrich mobile sensing datasets and support the development of more realistic, behaviorally grounded AI agents for simulation and decision-making.
\end{itemize}

\section{Related Work}
\textbf{Sociodemographic Inference Based on Human Mobility Patterns.} Recent studies leveraged machine learning methods to estimate sociodemographic attributes from mobility/activity features derived from trajectory data, often incorporating contextual signals such as semantic Points of Interest (POIs) \cite{zhong2015you, wu2019inferring} or social network information \cite{zhong2015you, chen2024coupling}. For example, \citet{chen2024coupling} predicted housing prices, used as a proxy for income, by combining mobility embeddings learned via Word2Vec with social network characteristics derived from call detail records (CDRs). Others have applied models such as XGBoost \cite{wu2019inferring}, Support Vector Machines (SVM) \cite{zhang2024social}, or CatBoost \cite{bakhtiari2023inferring} to infer attributes like gender, age, or income status from activity-derived or mobility-derived features. \citet{zhu2017prediction} predicts sociodemographic attributes from mobility trajectories by modeling variability in individual mobility patterns.

\textbf{LLMs for Modeling Cognitive Processes.} Recent works show that LLMs can simulate human cognitive processes through Theory-of-Mind (ToM) reasoning \cite{gandhi2023understanding, li2023theory, amirizaniani2024can, street2024llm}. To ensure alignment with human mental states, LLMs are most effective when supported by deliberate prompting \cite{gu2024simpletom} and structured reasoning frameworks \cite{zhou2023far}. Additionally, LLMs have demonstrated strong potential in inferring latent cognitive constructs \cite{ali2024comparing, chen2025perceptions} and predicting sociodemographic attributes from behavioral cues \cite{orlikowski2025beyond}, laying the foundation for cognitively grounded modeling of human behavior.

\textbf{Inverse Reinforcement Learning.} The potential of IRL to uncover latent intentions and inform downstream classification tasks has been demonstrated in \citet{hayes2011use, dadgostari2022identifying}. It has been increasingly applied to travel behavior analysis to uncover the underlying intentions. Methods include feature matching \cite{liu2022understanding}, maximum entropy \cite{koch2020review, okubo2024estimation}, and deep/adversarial IRL for high-dimensional settings \cite{zhao2023deep, liu2025personalized}. A recent work also exploreed model interpretability in mobility-focused IRL \cite{liang2025analyzing}.

To address IRL’s reward ambiguity \cite{ng2000algorithms}, some recent studies shifted from identifying a single solution to learning feasible reward sets \cite{metelli2021provably, metelli2023towards, zhao2023inverse}. Domain knowledge has long informed reward initialization \cite{liu2013understanding}, and more recently, LLMs have been used to guide both initialization and refinement \cite{ma2023eureka, kwon2023reward, chu2023accelerating}.

\begin{figure*}[!ht]
    \centering
    \includegraphics[width=0.8
    \textwidth]{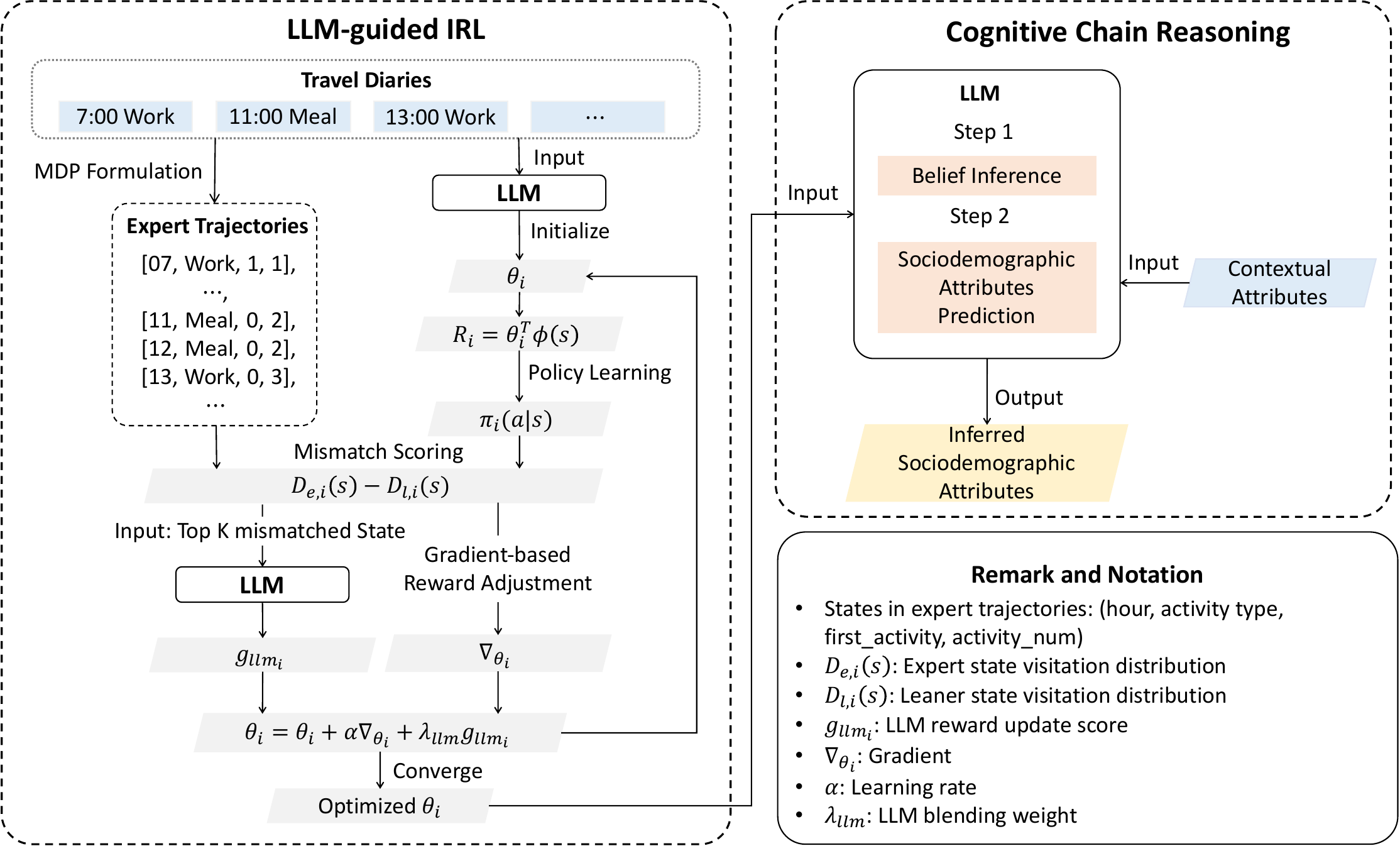}  
    \caption{Methodological Framework.}
    \label{fig:flow}
    \vspace{-1em}
\end{figure*}

\section{Methodology}
Our methodological framework is grounded in the \textbf{Theory of Planned Behavior (TPB)} (see Figure~\ref{fig:TPB}). TPB posits that background factors, such as sociodemographic and contextual attributes, influence people's beliefs, which include attitude, subjective norm, and perceived behavioral control. Attitude refers to an individual's evaluation of the behavior, subjective norm reflects perceived social pressure from peers or influential figures, and perceived behavioral control represents one's perceived ability to perform the behavior. These beliefs collectively inform intentions, and ultimately drive observed behaviors \cite{ajzen1991theory, ajzen2020theory}.

To infer individuals’ sociodemographic attributes from their observed mobility patterns, we propose to follow the TPB in reverse, and an overview of the proposed methodology is shown in Figure~\ref{fig:flow}. Specifically, individual trajectories are first converted into structured travel diaries that record detailed information on daily trips, such as activity types and departure times. These diaries are then represented as sequential activity patterns and modeled within a Markov Decision Process (MDP) framework. We subsequently introduce an LLM-guided inverse reinforcement learning (IRL) framework, in which the LLM provides heuristic guidance for reward initialization and updates to infer individuals’ latent intentions. Building on this, we propose a Cognitive Chain Reasoning (CCR) strategy that guides the LLM to first infer belief constructs from learned intentions and subsequently predict sociodemographic attributes using both these beliefs and external contextual variables (e.g., urbanicity, home location population density).


\subsection{MDP Framework}
We begin by modeling travel diaries as a Markov Decision Process (MDP), where each agent represents an individual making sequential travel decisions over time. Each dairy captures daily travel behavior, including features such as activity type and departure time. To enable structured temporal reasoning, we discretize the day into 24 hourly intervals, assuming one action decision is made at the start of each hour \cite{liang2025analyzing}. This is a reasonable assumption, as according to the National Household Travel Survey (NHTS), Americans made an average of approximately 3-4 trips per person per day \cite{NHTS2017}. 

An MDP is formally defined as $M = \{S, A, T, R, \gamma\}$, where $S$ denotes the set of states and $A$ the set of possible actions. The transition function $T(s, a, s{\prime})$ specifies the probability of transitioning to state $s{\prime} \in S$ after taking action $a \in A$ in state $s \in S$. The reward function 
$R(s)$ assigns a scalar reward to each state $s$, and the discount factor $\gamma \in [0, 1]$ controls the agent’s preference for immediate versus future rewards. Another key concept in the MDP framework is the policy $\pi(a|s)$, which defines the probability of an individual selecting action $a$ when in state $s$.

The five components of the MDP are specified as follows in the context of mobility modeling:
\paragraph{State Space \textbf{$S$}:} Each state $s \in S$ is represented as a feature vector $s = [h, a, f, n]$, where $h \in \{0, \cdots, 23\}$ denotes the hour of the day, $a \in \mathcal{A}$ indicates the activity type (i.e., home, work, education, escort and errand, leisure \cite{akar2012redefining}), $f \in \{0,1\}$ is a binary indicator of whether the activity is the first of the day, and $n \in R \geq 0$ captures the cumulative number of activities up to the current activity.
\paragraph{Action Space $A$:} We define $A = \{\text{\textbf{stay}}, \text{\textbf{travel}}\}$, where \textbf{stay} represents the decision to continue the current activity, and \textbf{travel} indicates a transition to a new activity.
\paragraph{Transition Probability Function $T$:} When $a = \text{\textbf{stay}}$, the transition is deterministic: the hour increments by one while all other state components remain unchanged, resulting in $s\prime = [h+1, a, f, n]$. In contrast, when $a = \text{\textbf{travel}}$, the transition is stochastic and modeled using empirical transition frequencies derived from the individual’s observed activity sequences. In this case, the hour also advanced by one $(h\prime = h + 1)$; the first-activity indicator is updated such that $f\prime = 0$ if $f = 1$, and remains unchanged otherwise; and the cumulative activity count increases by one $(n\prime = n + 1)$ as the activity changes.
\paragraph{Reward Function $R$:} The reward function $R(s)$ is defined as a linear function of the state features. Specifically, the reward assigned to a state $s$ is computed as:
\begin{equation}
R(s) = \boldsymbol{\theta}^\top \, \boldsymbol{\phi}(s)
\end{equation}
where $\boldsymbol{\phi}(s)$ denotes the concatenated feature vector comprising the one-hot hour indicators, one-hot activity type indicators, the binary first-activity indicator $f$, and the numerical feature $n$. The vector $\boldsymbol{\theta}$ contains the corresponding reward weight associated with each feature dimension.

\subsection{Intention Inference with LLM-guided IRL}
Once the MDP is formulated, IRL aims to infer individuals’ underlying intentions from observed mobility patterns. Specifically, it estimates reward weights such that the resulting policy produces trajectories, referred to as learner trajectories, that closely replicate the expert trajectories, i.e., the individuals’ observed state sequences.

We introduce an IRL framework in which LLMs provide guidance for both the initialization and iterative update of reward weights.

\underline{\textit{Reward Initialization:}} Inspired by \citet{ma2023eureka}, we use an LLM to initialize individual-specific reward weights given each individual’s multi-day travel diaries. The prompt template is provided in Figure~\ref{fig:reward-init-prompt} (Appendix \ref{sec:prompt}). In addition to interpreting the input diaries, the LLM leverages prior knowledge of human mobility behavior \cite{liu2024human} and inverse alignment capabilities \cite{sun2024inverse} to generate behaviorally meaningful and unique initializations. This mitigates the ill-posed nature of IRL \cite{ng2000algorithms, cao2021identifiability}, where multiple reward functions can explain the same observed behavior, and helps constrain the vast reward space \cite{adams2022survey} toward plausible solutions.

\underline{\textit{Policy Learning:}} Once the reward weights are initialized or updated, they are used to infer each individual’s policy $\pi_i(a|s)$. To account for the inherent variability and bounded rationality in human decision-making, we adopt a stochastic policy framework rather than assuming deterministic action selection. Individuals' behavior is modeled using the maximum entropy principle \citep{ziebart2008maximum}, wherein actions with higher expected value are assigned higher probabilities. Specifically, the value of a state $s$ is iteratively updated according to: 

{\small
\begin{equation}
V_i(s) = R_i(s) + log \sum_{a\in\mathcal{A}}\exp(\gamma \sum_{s\prime}T_i(s\prime|s,a)V_i(s\prime))
\end{equation}}

\noindent The updates proceed iteratively until the change in $V_i(s)$ across iterations falls below a predefined threshold  $\epsilon$. Upon convergence of $V_i(s)$, the corresponding stochastic policy $\pi_i(a|s)$ is computed as:
{\small
\begin{equation}
\pi_i(a|s) = \frac{\exp(R_i(s) + \gamma\sum_{s\prime}T_i(s\prime|s,a)V_i(s\prime))}{\sum_{a\prime}\exp(R_i(s) + \gamma\sum_{s\prime}T_i(s\prime|s,a\prime)V_i(s\prime))}
\end{equation}}

\underline{\textit{Mismatch Scoring:}} To iteratively refine reward weights, we compute a mismatch score as the discrepancy between learner and expert state visitation distributions. The expert state visitation distribution for individual $i$, denoted as $D_{e, i}(s)$, is estimated by counting the frequency of state visits across the individual's trajectories and normalizing the counts to obtain a probability distribution. The corresponding learner state visitation distribution, $D_{l,i}(s)$ is computed by simulating expected state occupancies under the learned policy $\pi_i(a|s)$ over a planning horizon of $\mathcal{T}(\mathcal{T}=24)$ steps. The initial state visitation distribution is defined by the empirical distribution of starting activities from expert trajectories. The learner’s state visitation distribution is then iteratively updated by propagating state occupancies according to the policy $\pi_i(a|s)$ and the transition dynamics $T_i(s\prime|s,a)$. At each time step $t$, the visitation distribution is updated as follows:

{\small
\begin{equation}
D_{l, i, t+1}(s\prime) = \sum_{s}\sum_{a}D_{l,i,t}(s) \cdot \pi_i(a|s) \cdot T_i(s\prime|s,a)
\end{equation}}

\noindent The top $K$ states exhibiting the largest absolute mismatch score $|D_{e,i}(s)-D_{l,i}(s)|$ between expert and learner visitation frequencies are identified to use as inputs for the LLM-guided reward update procedure within the iterative learning process.

\underline{\textit{LLM-guided Reward Updates:}}
In the early stages of training, the convergence of reward weights may be suboptimal, particularly given the bias introduced by traditional Maximum Entropy gradient ascent methods when applied to sparse expert trajectories, such as multi-day activity sequences. To address this issue, an LLM is incorporated to provide heuristic reward weight adjustments during updates. Leveraging the LLM’s prior knowledge of activity behavior patterns further mitigates the challenges posed by data sparsity and improves the robustness of the learning process. Specifically, the reward weight vector $\theta_i$ for individual $i$ is iteratively updated according to:

{\small
\begin{equation}
\nabla_{\theta_i} = \frac{1}{n_i}\sum_{j=1}^{n_i}\sum_{s\in\tau_{i,j}}\phi(s) - \frac{1}{\mathcal{T}}\sum_{t=0}^{\mathcal{T}-1}\sum_{s}D_{l,i,t}(s)\cdot\phi(s)
\end{equation}

\begin{equation}
\Delta \theta_i = \alpha \cdot \nabla_{\theta_i} + \lambda_{\text{LLM}} \cdot \mathbf{g}_{\text{LLM}_i}
\end{equation}

\begin{equation}
\theta_i = \theta_i + \Delta \theta_i
\end{equation}}

\noindent Here, $\nabla_{\theta_i}$ denotes the Maximum Entropy IRL gradient, $n_i$ is the number of expert trajectories for individual $i$, and $\tau_{i,j}$ represents the $j$-th expert trajectory of individual $i$. The parameter $\alpha$ is the learning rate, $\lambda_{\text{LLM}}$ is a tunable blending weight, and $\mathbf{g}_{\text{LLM}}$ denotes the LLM-based update direction. 

For LLM-guided reward updates, the top-$K$ states with the highest absolute mismatch scores are provided as input to the LLM. The LLM then suggests an update direction for each of the reward weights. The LLM’s output for each weight is restricted to one of three discrete values: {-1, 0, 1}, corresponding to a decrease, no change, or an increase. The complete prompt template used for this process is presented in Figure \ref{fig:reward-update-prompt} in Appendix \ref{sec:prompt}. The updates continue iteratively until convergence, which is determined by either the change in $\theta_i$ or the KL divergence between the expert and learner state visitation distributions dropping below a predefined threshold $\epsilon’$. The KL divergence formulation is detailed in Appendix \ref{subsec: KL}.

\subsection{Sociodemographic Attribute Prediction with Cognitive Chain Reasoning}
After inferring individual-specific reward weights that represent latent intentions, we leverage these to predict individual-level sociodemographic attributes. Grounded in the TPB, sociodemographic and contextual attributes act as background variables that indirectly shape intentions through their influence on beliefs. Accordingly, we first infer latent beliefs from the learned reward weights and then predict sociodemographic attributes by integrating these beliefs with external contextual features.

To implement this approach, we design a CCR prompting template that guides the LLM to sequentially reason through three beliefs before making a prediction. This structure aligns the model’s reasoning process with the TPB to enhance its theoretical consistency. Furthermore, the CCR framework capitalizes on the LLM’s strengths in multi-step reasoning capabilities that are particularly important for mapping abstract reward weights into psychologically meaningful beliefs and, ultimately, individuals' sociodemographic profiles. Contextual attributes are also incorporated to aid sociodemographic prediction following belief inference, as they jointly influence belief formation alongside sociodemographic attributes. This accounts for the fact that similar belief patterns may correspond to different sociodemographic groups under varying contextual conditions. The CCR prompting template is provided in Figure \ref{fig: ccr} (Appendix \ref{sec:prompt}).

\section{Experiment}
\subsection{Experiment Setup}
\noindent \textbf{Implementation Details.} Please find the implementation details in Appendix \ref{subsec:implement}.

\noindent \textbf{Baselines.}
We evaluate our proposed model against a set of baseline models for predicting sociodemographic attributes from mobility data, including SVM, XGBoost, CatBoost, and GPT-4o. The first three models are selected based on their demonstrated effectiveness in prior work \cite{ zhu2017prediction, wu2019inferring, bakhtiari2023inferring}. In addition, we include GPT-4o as a baseline to explore the emerging capability of LLMs to infer sociodemographic attributes directly from mobility patterns in a zero-shot setting. 

We evaluate the models using either extracted mobility features or direct travel diary data, both combined with contextual attributes. Specifically, extracted features, selected based on prior literature, capture key aspects of individual mobility behavior, such as averages and variations in departure times, activity types, trip distances, travel times, and trip frequencies. The extracted features are detailed in Table \ref{tab:extracted_feature} in Appendix \ref{subsec: extract}. To reduce overfitting and ensure fair evaluation, feature selection is performed on the training set (Appendix~\ref{subsec:feature}). The same set of selected features is also provided as input to GPT-4o to infer sociodemographic attributes from the test set. Direct travel diary data, containing multiple-day trip records, along with the contextual data, are converted into semantically descriptive inputs for GPT-4o. This setup is designed to evaluate GPT-4o’s ability to directly infer sociodemographic attributes from mobility data without relying on feature engineering.

\noindent \textbf{Datasets.} To obtain both individuals' travel trajectories and their sociodemographic profiles, we use the 2017 Puget Sound Regional Council Household Travel Survey \cite{nrel_tsdc}. It comprises data from 6,254 participants, covering a total of 52,492 trips. It includes two mobility components: a one-day household travel diary and a seven-day smartphone-based GPS diary. In our study, we only used the GPS diaries for analysis as they are derived from passive collected multi-day trajectories that better reflect naturalistic behavior. The GPS diaries provide detailed individual-level trip and activity information, such as travel day, departure and arrival times, activity type, and trip distance, which can be readily inferred from any raw GPS trajectory data.

To ensure data quality and analytical relevance, we applied a series of filtering criteria to the GPS diary data (Appendix~\ref{subsec: data_selection}). After applying the filtering criteria, the final dataset consists of 617 qualified individuals, contributing a total of 11,964 trips for analysis. Additionally, to support contextualized reasoning in prediction, we incorporated contextual variables as auxiliary inputs. Details of contextual attributes are provided Appendix~\ref{subsec: context}.

\noindent \textbf{Evaluation Metrics.}
We use KL divergence and L1 distance to quantify the discrepancy between expert and learner policies. To evaluate prediction performance, we reporte class-level Precision, Recall, and F1 score, as well as overall Accuracy and Weighted F1 score. Metric definitions are in Appendix~\ref{subsec: KL} and~\ref{subsec:classification}.

\subsection{Main Results}
We evaluate our model on gender and age prediction (see Appendix~\ref{sec: income} for income and employment), comparing it with baselines to assess the effectiveness of incorporating psychological theory and LLM prior knowledge without relying on additional training data.

\begin{table}[h]
\scriptsize
\centering
\setlength{\tabcolsep}{2.5pt}
\caption{
Model comparison for gender prediction. The proposed model outperforms all baselines in both overall and class-level performance.}
\vspace{-1em}
\begin{tabular}{llccccc}
\toprule
\textbf{Method} & \textbf{Class} & \textbf{Precision} & \textbf{Recall} & \textbf{F1-score} & \textbf{Accuracy} & \textbf{Weighted F1} \\ 
\midrule

\multirow{2.6}{*}{\begin{tabular}[c]{@{}l@{}}SVM\end{tabular}} 
  & Male   & 0.603 & 0.633 & 0.618 
  & \multirow[c]{2.6}{*}{0.621} & \multirow[c]{2.6}{*}{0.621} \\ 
\cmidrule(lr){3-5}
  & Female & 0.639 & 0.609 & 0.624 
  &                           &                           \\ 
\midrule

\multirow[c]{2.6}{*}{\begin{tabular}[c]{@{}l@{}}XGBoost\end{tabular}} 
  & Male   & 0.597 & 0.667 & 0.630 
  & \multirow[c]{2.6}{*}{0.621} & \multirow[c]{2.6}{*}{0.620} \\ 
\cmidrule(lr){3-5}
  & Female & 0.649 & 0.578 & 0.612 
  &                           &                           \\ 
\midrule

\multirow[c]{2.6}{*}{\begin{tabular}[c]{@{}l@{}}CatBoost\end{tabular}} 
  & Male   & 0.683 & 0.642 & 0.662 
  & \multirow[c]{2.6}{*}{0.645} & \multirow[c]{2.6}{*}{0.646} \\ 
\cmidrule(lr){3-5}
  & Female & 0.607 & 0.649 & 0.627 
  &                           &                           \\ 
\midrule

\multirow[c]{2.6}{*}{\begin{tabular}[c]{@{}l@{}}GPT-4o\end{tabular}} 
  & Male   & 0.627 & 0.700 & 0.661 
  & \multirow[c]{2.6}{*}{0.653} & \multirow[c]{2.6}{*}{0.653} \\ 
\cmidrule(lr){3-5}
  & Female & 0.684 & 0.609 & 0.645 
  &                           &                           \\ 
\midrule

\multirow[c]{2.6}{*}{\begin{tabular}[c]{@{}l@{}}GPT-4o \\ (Diaries)\end{tabular}} 
  & Male   & 0.625 & 0.667 & 0.645 
  & \multirow[c]{2.6}{*}{0.645} & \multirow[c]{2.6}{*}{0.645} \\ 
\cmidrule(lr){3-5}
  & Female & 0.667 & 0.625 & 0.645 
  &                           &                           \\ 
\midrule

\multirow[c]{2.6}{*}{\begin{tabular}[c]{@{}l@{}}\textit{SILIC}\end{tabular}} 
  & Male   & 0.920 & 0.767 & 0.836 
  & \multirow[c]{2.6}{*}{\textbf{0.855}} & \multirow[c]{2.6}{*}{\textbf{0.853}} \\ 
\cmidrule(lr){3-5}
  & Female & 0.811 & 0.938 & 0.870
  &                                   &                                   \\ 
\bottomrule
\end{tabular}
\label{tab: gender}
\end{table}

As shown in Table \ref{tab: gender}, GPT-4o slightly outperforms traditional machine learning models using the same input features, demonstrating its ability to distinguish behavioral patterns through prior domain knowledge. Moreover, its comparable performance with direct travel diaries suggests that it can operate effectively without relying on explicit feature engineering. Our proposed model further advances GPT-4o’s inference from observed mobility data, yielding over a 30\% improvement in overall accuracy. It also achieves more balanced performance across classes, as evidenced by higher weighted and class-specific F1 scores. 

\begin{table}[h]
\scriptsize
\centering
\setlength{\tabcolsep}{2.5pt}
\caption{
Model comparison for age prediction. The proposed model outperforms all baselines in both overall and class-level performance, and effectively identifies age groups that baseline models struggle to classify. }
\vspace{-1em}
\begin{tabular}{llccccc}
\toprule
\textbf{Method} & \textbf{Class} & \textbf{Precision} & \textbf{Recall} & \textbf{F1-score} & \textbf{Accuracy} & \textbf{Weighted F1} \\ 
\midrule

\multirow[c]{4.3}{*}{\begin{tabular}[c]{@{}l@{}}SVM\end{tabular}} 
  & 18--44 & 0.752 & 0.989 & 0.854 
  & \multirow[c]{4.3}{*}{0.742} & \multirow[c]{4.3}{*}{0.644} \\ 
\cmidrule(lr){3-5}
  & 45--64 & 0.000 & 0.000 & 0.000 
  &                            &                            \\ 
\cmidrule(lr){3-5}
  & 65+    & 0.333 & 0.200 & 0.250 
  &                            &                            \\ 
\midrule

\multirow[c]{4.3}{*}{\begin{tabular}[c]{@{}l@{}}XGBoost\end{tabular}} 
  & 18--44 & 0.767 & 0.859 & 0.810 
  & \multirow[c]{4.3}{*}{0.677} & \multirow[c]{4.3}{*}{0.650} \\ 
\cmidrule(lr){3-5}
  & 45--64 & 0.250 & 0.148 & 0.186 
  &                            &                            \\ 
\cmidrule(lr){3-5}
  & 65+    & 0.200 & 0.200 & 0.200 
  &                            &                            \\ 
\midrule

\multirow[c]{4.3}{*}{\begin{tabular}[c]{@{}l@{}}CatBoost\end{tabular}} 
  & 18--44 & 0.792 & 0.870 & 0.829 
  & \multirow[c]{4.3}{*}{0.718} & \multirow[c]{4.3}{*}{0.700} \\ 
\cmidrule(lr){3-5}
  & 45--64 & 0.421 & 0.296 & 0.348 
  &                            &                            \\ 
\cmidrule(lr){3-5}
  & 65+    & 0.250 & 0.200 & 0.222 
  &                            &                            \\ 
\midrule

\multirow[c]{4.3}{*}{\begin{tabular}[c]{@{}l@{}}GPT-4o\end{tabular}} 
  & 18--44 & 0.870 & 0.870 & 0.870 
  & \multirow[c]{4.3}{*}{0.710} & \multirow[c]{4.3}{*}{0.732} \\ 
\cmidrule(lr){3-5}
  & 45--64 & 0.615 & 0.296 & 0.400 
  &                            &                            \\ 
\cmidrule(lr){3-5}
  & 65+    & 0.000 & 0.000 & 0.000 
  &                            &                            \\ 
\midrule

\multirow[c]{4.3}{*}{\begin{tabular}[c]{@{}l@{}}GPT-4o\\(Diaries)\end{tabular}} 
  & 18--44 & 0.863 & 0.891 & 0.877 
  & \multirow[c]{4.3}{*}{0.734} & \multirow[c]{4.3}{*}{0.740} \\ 
\cmidrule(lr){3-5}
  & 45--64 & 0.571 & 0.296 & 0.390 
  &                            &                            \\ 
\cmidrule(lr){3-5}
  & 65+    & 0.067 & 0.200 & 0.100 
  &                            &                            \\ 
\midrule

\multirow[c]{4.3}{*}{\begin{tabular}[c]{@{}l@{}}\textit{SILIC}\end{tabular}} 
  & 18--44 & 0.887 & 0.989 & 0.935
  & \multirow[c]{4.3}{*}{\textbf{0.863}} & \multirow[c]{4.3}{*}{\textbf{0.844}} \\ 
\cmidrule(lr){3-5}
  & 45--64 & 0.900 & 0.375 & 0.529
  &                                   &                                   \\ 
\cmidrule(lr){3-5}
  & 65+    & 0.500 & 0.800 & 0.615
  &                                   &                                   \\ 
\bottomrule
\end{tabular}
\label{tab: age}
\end{table}

Table \ref{tab: age} presents the age prediction results for a multi-class task. While baseline models achieve decent overall accuracy, they struggle to identify individuals over the age of 45, particularly those over 65, as evidenced by their low class-level F1 scores. For machine learning models, this performance gap may be attributed to data imbalance, as individuals aged 45 and above, particularly those over 65, are underrepresented in the training set. Consequently, they are biased toward the majority class, resulting in diminished predictive performance for those age groups. The GPT 4o inference model, despite its general reasoning capabilities, may have difficulty distinguishing age-specific behavioral patterns in mobility without access to the underlying cognitive patterns. In contrast, our proposed model achieves superior performance compared to all baseline methods, with significantly higher overall prediction accuracy and weighted F1 score, while also demonstrating improved ability to distinguish adults across different age groups. This improvement is particularly important for downstream applications that rely on accurately identifying targeted sociodemographic groups.

Overall, the proposed model consistently achieves strong performance in both total prediction accuracy and class-level identification across all sociodemographic attributes. These results highlight the effectiveness of inversely following the well-established behavioral theory by capturing mediating cognitive processes for inferring sociodemographic characteristics from mobility patterns.

\subsection{Ablation Study}
We conducted ablation experiments using the GPT-4o model to evaluate the effectiveness of (1) LLM-guided reward weight initialization and updates and (2) the CCR strategy. 

First, we compare the KL divergence and L1 distance (definitions in Appendix \ref{subsec: KL}) between expert and learner behaviors under different settings to evaluate whether the policy derived from the final reward weights effectively replicates expert trajectories. As shown in Table \ref{tab: ILR abla}, removing both LLM-guided initialization and updates results in over a 41\% increase in KL divergence and over a 10\% increase in L1 distance, confirming their contribution. When only the LLM-guided updates are replaced with standard gradient ascent, the increase is less pronounced, highlighting the importance of LLM-guided initialization in constraining the reward space effectively. Furthermore, retaining only the LLM-guided reward weight updates still achieves better performance than removing both the initialization and update steps, indicating the effectiveness of LLM-provided heuristics in guiding optimization toward behaviorally plausible reward structures.

\begin{table}[!h]
\scriptsize
\centering
\setlength{\tabcolsep}{1.7pt}
\caption{\label{citation-guide}
Ablation Study of LLM-guided IRL. It demonstrates the effectiveness of both LLM-based reward initialization and iterative update guidance.}
\vspace{-1em}
\begin{tabular}{lcc}
\toprule
 & KL Divergence & L1 Distance \\ \midrule
Random Initialization + Gradient Ascent & 0.594 & 0.722  \\
LLM-guided Initialization + Gradient Ascent & 0.447 & 0.696 \\
Random Initialization + LLM-guided updates & 0.543 & 0.704 \\
LLM-guided Initialization + updates & \textbf{0.419} & \textbf{0.654} \\ \bottomrule
\end{tabular}
\label{tab: ILR abla}
\end{table}

Second, we compare the Accuracy and Weighted F1 of our proposed model against two variants that substitute the CCR module with either pure GPT-4o inference or Chain-of-Thought (CoT) inference, in order to assess the effectiveness of the CCR module. As shown in Table \ref{tab: CCR}, the CoT strategy outperforms pure inference by leveraging the structured reasoning capabilities of the LLM. By further guiding the reasoning process with the well-established behavioral theory, the CCR module aligns more closely with human cognitive processes, leading to the best performance among all compared methods.

\begin{table}[!h]
\scriptsize
\centering
\setlength{\tabcolsep}{2.5pt}
\caption{\label{tab:ablation-pred}
Ablation study of CCR on gender and age prediction. CCR outperforms both pure inference and CoT.
}
\vspace{-0.8em}
\begin{tabular}{lcccc}
\toprule
\multirow{2.6}{*}{Method} & \multicolumn{2}{c}{\textbf{Gender}} & \multicolumn{2}{c}{\textbf{Age}} \\
\cmidrule(lr){2-3} \cmidrule(lr){4-5}
 & \textbf{Accuracy} & \textbf{Weighted F1} & \textbf{Accuracy} & \textbf{Weighted F1} \\
\midrule
IRL + Inference & 0.718 & 0.708 & 0.815 & 0.797 \\
IRL + CoT & 0.766 & 0.766 & 0.823 & 0.812 \\
IRL + CCR & \textbf{0.855} & \textbf{0.853} & \textbf{0.863} & \textbf{0.844} \\
\bottomrule
\end{tabular}
\label{tab: CCR}
\end{table}

\section{Conclusion}
This study proposes \textit{SILIC}, a novel IRL+CCR framework to predict sociodemographic attributes from mobility patterns by inversely aligning with the TPB. We leverage IRL to infer individual-specific latent intentions, while addressing its key methodological challenges through heuristic guidance provided by LLMs. Building on inferred intentions, the CCR module guides LLM reasoning to sequentially infer belief constructs and predict sociodemographic attributes. Contextual features are incorporated to support informed predictions, as sociodemographic and contextual factors jointly influence behavioral beliefs. Extensive experiments across multiple prediction tasks demonstrate that our framework not only significantly outperforms established baselines, but also offers a solution for enriching large-scale real-world or synthetic trajectory datasets, with implications for various applications involving human behavior modeling.

\section*{Limitations}

Although the proposed model shows strong potential for inferring sociodemographic attributes from trajectories, several limitations remain. First, while the model infers latent intentions and belief constructs from observed behavior, these internal variables lack ground truth for direct validation. Any misalignment in these inferred representations may propagate to downstream sociodemographic predictions. Second, the IRL component depends on heuristic support from LLMs, whose guidance—partially derived from general domain knowledge—may fail to fully capture behavioral nuances specific to certain geographic regions. Third, the current state space design may omit relevant mobility-related features (e.g., trip distance), limiting its capacity to fully represent human decision-making dynamics. Future work should address these challenges. For example, leveraging survey data that include cognitive or attitudinal variables to assess the validity of inferred intentions and beliefs, and incorporating human-in-the-loop calibration to refine LLM-generated reward priors or updates to better reflect regional characteristics.

\section*{Acknowledgment}
This research was supported by the National Science Foundation (Award \#2338959 and \#2416202). Any opinions, findings, conclusions, or recommendations expressed in this material are those of the authors and do not necessarily reflect the views of NSF.

\bibliography{custom}

\clearpage
\appendix
\label{sec:appendix}
\section{Implementation Details}

\subsection{Implementation Details}
\label{subsec:implement}
We implement our full framework using GPT-4o as the backbone for both the LLM-guided IRL module and the CCR module. The data is split into 80\% training and 20\% testing. Evaluation is performed on the held-out test set. We set the hyperparameters as follows: $K$ = 30, $\alpha = 2$, $\lambda_{\text{LLM}} = 0.002$, and convergence thresholds $\epsilon = \epsilon’ = 10^{-4}$. 

\subsection{Activity Mapping}
\label{sebsec: activity}
To reduce the state space, we group activities into five high-level categories according to the following mapping scheme:
\begin{itemize}
\item \textbf{Home}: Home
\item \textbf{Work}: Work, Work-related
\item \textbf{Education:} School
\item \textbf{Escort and Errand: }Personal Business / Errand / Appointment, Escort, Change Model
\item \textbf{Leisure:} Social / Recreational, Shopping, Meal, Other
\end{itemize}

\subsection{Feature Selection}
\label{subsec:feature}
\begin{figure}[h]
    \centering
    \includegraphics[width=0.45
    \textwidth]{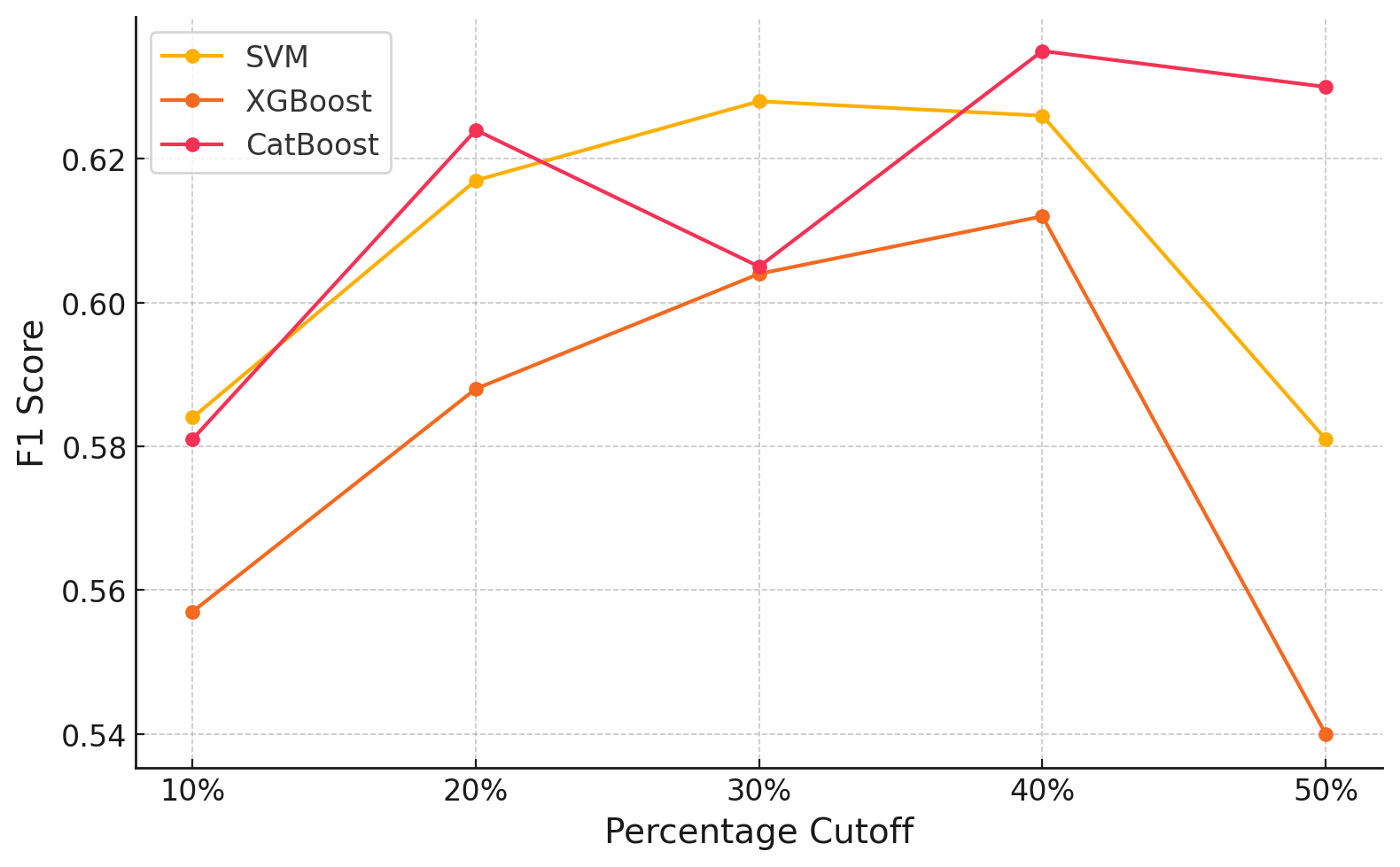}
    \vspace{-0.5em}
    \caption{F1 score variation across feature selection thresholds for gender prediction using SVM, XGBoost, and CatBoost.}
    \label{fig:f1_gender}
    \vspace{-1em}
\end{figure}

\begin{figure}[h]
    \centering
    \includegraphics[width=0.45
    \textwidth]{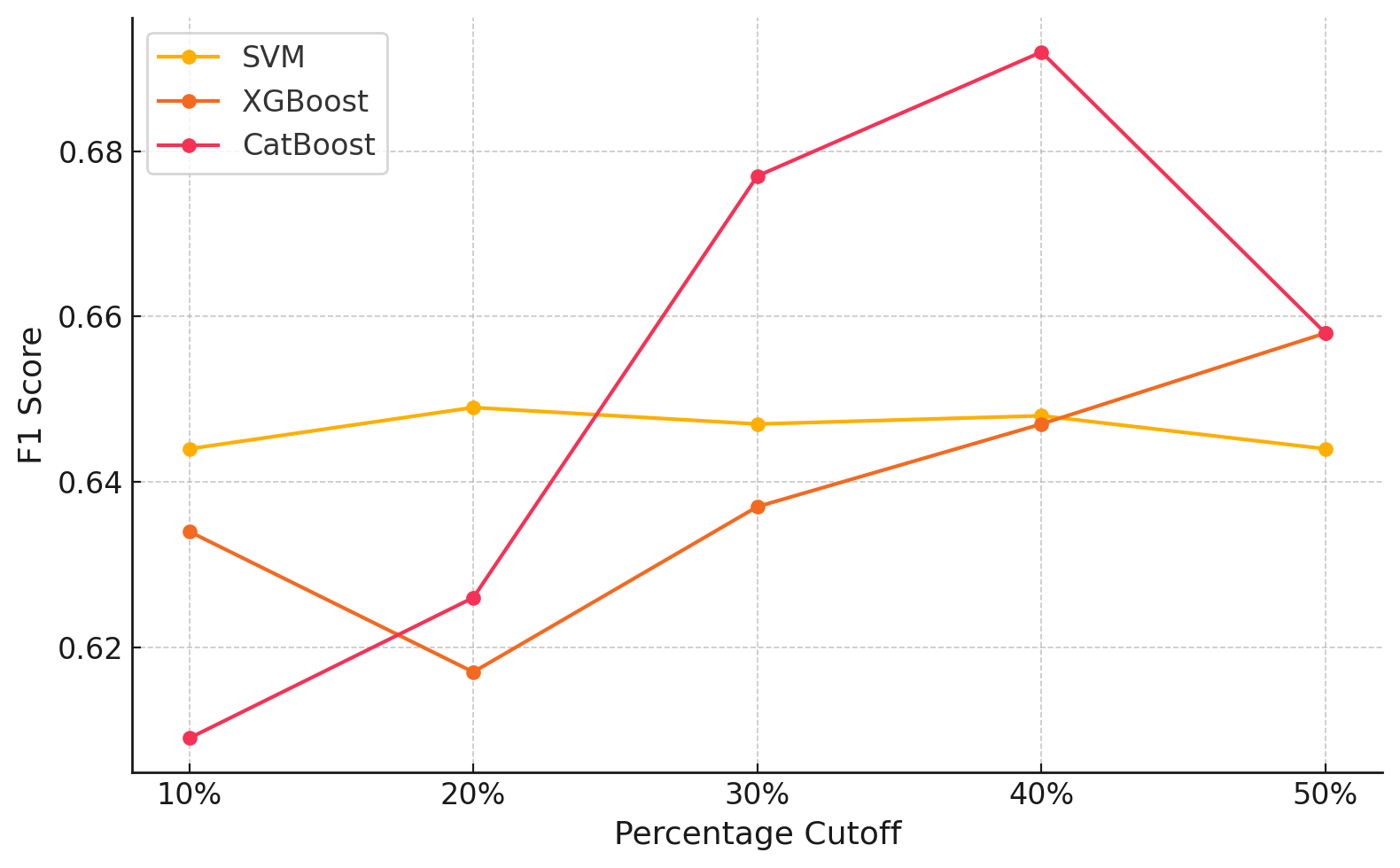}
    \vspace{-0.5em}
    \caption{F1 score variation across feature selection thresholds for age prediction using SVM, XGBoost, and CatBoost.}
    \label{fig:f1_age}
    \vspace{-1em}
\end{figure}

We compute ANOVA F-scores \cite{shakeela2021optimal} to assess the statistical association between each feature and the target variable in the training set, rank the features accordingly, and retain only those within the top 40\%, corresponding to scores above the 60th percentile threshold. This selection aims to optimize the performance of baseline machine learning models, ensuring a fair comparison with our proposed approach. This filtering process is applied to the combined set of extracted and contextual attributes. Figures \ref{fig:f1_gender} and \ref{fig:f1_age} present line plots illustrating how F1 scores vary with different feature selection thresholds across the three baseline machine learning models for gender and age prediction, respectively, using five-fold cross-validation.

\subsection{Data Selection}
\label{subsec: data_selection}
We applied data selection procedures to ensure both the quality and relevance of the dataset for our analysis. First, we retained only the household representatives (defined as the primary survey respondent \cite{nrel_tsdc}), as they typically provide the most complete and reliable information within each household. Next, we excluded individuals who did not complete the survey to ensure the completeness of sociodemographic labels and mobility records. We focused on weekday trips to capture routine mobility behavior, which is more informative for modeling stable mobility patterns. Finally, we retained only individuals with at least two days of GPS trajectory data to ensure sufficient behavioral observations for effective IRL model training.

\subsection{Evaluation Metrics for IRL}
\label{subsec: KL}
To quantify the discrepancy between expert and learner policies, we adopt both KL divergence and L1 distance. For each individual $i$, KL divergence captures how much the learned policy $\pi_{l,i}$ deviates from the expert policy $\pi_{e,i}$, weighted by the empirical state visitation distribution. L1 distance computes the average absolute difference between $\pi_{l,i}$ and $\pi_{e,i}$ across states. The formal definitions of KL divergence and L1 distance are presented as follows:

{\small
\begin{equation}
KL(\pi_{e,i}||\pi_{l,i}) = \sum_sD_{e,i}(s)\sum_a\pi_{e,i}(a|s)\log(\frac{\pi_{e,i}(a|s)}{\pi_{l,i}(a|s)})
\end{equation}}

{\small
\begin{equation}
L1(\pi_{e,i}, \pi_{l,i}) = \frac{1}{|S|}\sum_{s\in S}\sum_a|\pi_{e,i}(a|s)-\pi_{l,i}(a|s)|
\end{equation}}

\subsection{Evaluation Metrics for Classification Tasks}
\label{subsec:classification}
To assess the effectiveness of the proposed model in predicting sociodemographic attributes from observed mobility patterns, we evaluate its performance against ground truth labels using standard classification metrics: class-level \textbf{Precision}, \textbf{Recall} and \textbf{F1-score}, and total-level \textbf{Accuracy} and \textbf{Weighted F1-score}.

Specifically, for each class $c \in C$, let $TP_c$, $TN_c$, $FP_c$, and $FN_c$ represent the number of true positives, true negatives, false positives, and false negatives, respectively. Let $N_c$ denote the total number of instances belonging to class $c$ (i.e., the support), and let $N$ denote the total number of instances across all classes. Based on these definitions, the evaluation metrics are defined as follows:
\begin{equation}
Precision_c = \frac{TP_c}{TP_c+FP_c}
\end{equation}
\begin{equation}
Recall_c = \frac{TP_c}{TP_c+FN_c}
\end{equation}
\begin{equation}
F1_c = \frac{2\cdot Precision_c\cdot Recall_c}{Prevision_c+Recall_c}
\end{equation}
\begin{equation}
Accuracy = \frac{\sum_{c\in C} TP_c + TN_c}{\sum_{c\in C}TP_c+TN_c+FP_c+FN_c}
\end{equation}
\begin{equation}
Weighted \ F1 = \sum_{c\in C} \frac{N_c}{N}\cdot F1_c
\end{equation}

\onecolumn
\newpage
\section{Prompt Design}
\label{sec:prompt}
\begin{figure}[H]
\noindent\rule{\textwidth}{1pt}

\vspace{0.5em}
\noindent\textbf{Prompt Template for Reward Initialization}
\\
\noindent\rule{\textwidth}{0.5pt}

\vspace{0.8em}

\noindent\textbf{Task Description} \\
You are an expert in travel behavior modeling and inverse reinforcement learning (IRL). You are provided with an individual’s multi-day weekday travel diaries, presented as a chronological sequence of activities and their corresponding departure times.

\vspace{0.5em}

\noindent\textbf{Input:}
\begin{itemize}
    \item \textbf{Travel Diaries:} \textit{[Travel Diaries]}
    \item \textbf{State Features:} $\boldsymbol{\phi}(s)$, a feature vector encoding relevant state attributes.
\end{itemize}

\vspace{0.5em}

\noindent\textbf{Objective} \\
Estimate an initial reward weight vector $\boldsymbol{\theta}$ for an IRL model, where the reward function is defined as:
\[
R(s) = \boldsymbol{\theta}^\top \cdot \boldsymbol{\phi}(s)
\]

\vspace{0.5em}

\noindent\textbf{Instructions:}
\begin{itemize}
    \item Analyze the provided travel diaries to identify patterns in the individual’s observed activity behavior.
    \item Apply general travel behavior knowledge (e.g., typical preferences for activity types and times of day).
    \item Assign meaningful weights to the corresponding features in $\boldsymbol{\phi}(s)$.
    \item For features with insufficient or ambiguous evidence, assign values near zero.
\end{itemize}

\vspace{0.5em}

\noindent\textbf{Output Format} \\
Return the 31-dimensional vector $\boldsymbol{\theta}$ as a valid Python list of 31 float values. Each value must be between $-2$ and $2$. No additional explanation or formatting should be included.

\vspace{0.5em}

\noindent\rule{\textwidth}{0.5pt}

\caption{Prompt used to initialize reward weights for IRL}
\label{fig:reward-init-prompt}
\end{figure}

\newpage

\begin{figure}[H]
\noindent\rule{\textwidth}{1pt}

\vspace{0.5em}

\textbf{Prompt Template for Reward Weight Update}

\noindent\rule{\textwidth}{0.5pt}

\vspace{0.8em}

\noindent\textbf{Task Description} \\
You are assisting in tuning the reward weight vector $\boldsymbol{\theta}$ for a maximum entropy inverse reinforcement learning (IRL) model. The objective is to adjust $\boldsymbol{\theta}$ to better align the learner’s state visitation distribution with that of the expert.

\vspace{0.5em}

\noindent\textbf{Input:}
\begin{itemize}
    \item \textbf{Current Reward Weights:} $\boldsymbol{\theta}$ = \textit{[current $\theta$ values]}
    \item \textbf{State Mismatches:} A list of the top 30 state mismatches, where each state is represented as a 4-tuple $(\text{hour}, \text{activity\_type}, \text{is\_first\_trip}, \text{activity\_segment\_count})$, along with the expert and learner visitation frequencies for each state.
\end{itemize}

\vspace{0.5em}

\noindent\textbf{Reward Function} \\
The reward function is defined as:
\[
R(s) = \boldsymbol{\theta}^\top \cdot \boldsymbol{\phi}(s)
\]
where $\boldsymbol{\phi}(s)$ is a 31-dimensional feature vector:
\begin{itemize}
    \item 5 one-hot indicators for activity type: [Home, Work, School, Errand and Escort, Leisure]
    \item 24 hour-of-day indicators: [hour\_0 to hour\_23]
    \item 1 binary indicator: is\_first\_trip
    \item 1 normalized numeric feature: activity\_segment\_count
\end{itemize}

\vspace{0.5em}

\noindent\textbf{Objective} \\
Based on the provided state mismatches and your prior domain knowledge, suggest an \textbf{update direction} for each of the 31 reward weights to reduce the discrepancies between the expert and learner state visitation distributions.

\vspace{0.5em}

\noindent\textbf{Instructions:}
\begin{itemize}
    \item Analyze the provided state mismatches and determine whether each feature weight in $\boldsymbol{\theta}$ should be increased, decreased, or left unchanged.
    \item For each of the 31 reward weights, output an update direction constrained to \{-1, 0, 1\}, where -1 indicates decrease, 0 indicates no change, and 1 indicates increase.
\end{itemize}

\vspace{0.5em}

\noindent\textbf{Output Format} \\
Return a Python list of 31 integers, each being -1, 0, or 1. No additional text or explanation should be included.

\vspace{0.5em}

\noindent\rule{\textwidth}{0.5pt}

\caption{Prompt template used to generate reward update directions from the LLM.}
\label{fig:reward-update-prompt}
\end{figure}

\begin{figure}[H]
\noindent\rule{\textwidth}{1pt}

\vspace{0.5em}
\noindent\textbf{Prompt Template for Sociodemographic Attribute Prediction with CCR}
\\
\noindent\rule{\textwidth}{0.5pt}

\vspace{0.8em}

\noindent\textbf{Task Description} \\
You are an expert in travel behavior modeling. In this task, you will apply the Theory of Planned Behavior (TPB) in reverse to infer \textbf{\emph{[age} / \emph{gender} / \emph{income level} / \emph{employment status]}} from the individual’s intention, as captured by the learned reward weights, and environmental context.

\vspace{0.5em}

\noindent\textbf{Input:}
\begin{itemize}
    \item \textbf{Reward Weights:} $\boldsymbol{\theta}$, a 31-dimensional vector representing the individual’s inferred preferences over activities, time-of-day, and trip structure.
    \item \textbf{Environmental Context:} External attributes such as urban/rural, population density, distance to transit, and housing characteristics.
\end{itemize}

\vspace{0.5em}

\noindent\textbf{Objective} \\
Predict the individual's \textbf{\emph{[age} / \emph{gender} / \emph{income level} / \emph{employment status]}} by interpreting the beliefs encoded in $\boldsymbol{\theta}$, supported by the environmental context.

\vspace{0.5em}

\noindent\textbf{Instructions:}
\begin{enumerate}
    \item \textbf{Step 1: Belief Inference} \\
    Analyze the reward weights to identify the individual's underlying attitude, subjective norm, and perceived behavioral control.

    \item \textbf{Step 2: Sociodemographic Prediction} \\
    Combine the inferred beliefs with the provided environmental context to predict the individual's \textbf{\emph{[age} / \emph{gender} / \emph{income level} / \emph{employment status]}}.
\end{enumerate}

\noindent Follow the above two steps in order when generating your prediction.

\vspace{0.5em}

\noindent\textbf{Output Format} \\
Return only the predicted label (e.g., \texttt{0}, \texttt{1}, or \texttt{2}) corresponding to the target category. No explanation or additional formatting should be included.

\vspace{0.5em}
\noindent\rule{\textwidth}{0.5pt}
\caption{CCR prompt for predicting sociodemographic attributes from IRL reward weights and contextual attributes}
\label{fig: ccr}
\end{figure}
\clearpage

\twocolumn

\section{Contextual and Extracted Features}
\subsection{Extracted Features}
\label{subsec: extract}

\begin{table}[ht]
\captionsetup{font=normalsize}
\normalsize
\centering
\caption{Extracted Features and Definitions.}
\label{tab:extracted_feature}
\vspace{0.5em}
\begin{tabular}{p{5cm} p{9.5cm}}
\toprule
\textbf{Extracted Feature} & \textbf{Definition} \\
\midrule
Trip Distance & Average distance per trip (in miles) \\
Std Trip Distance & Standard deviation of trip distance (in miles) \\
Num Trips per Day & Average number of trips taken daily \\
Std Num Trips per Day & Standard deviation of daily trip counts \\
Destination Entropy & Diversity of unique destination locations visited \\
\% Work Trips & Percentage of trips for work or work-related purposes \\
\% School Trips & Percentage of trips for school or educational purposes \\
\% Shopping Trips & Percentage of trips for shopping activities \\
\% Social/Recreation Trips & Percentage of trips for social or recreational purposes \\
\% Errand Trips & Percentage of trips for business, errands, or appointments \\
\% Escort Trips & Percentage of escort-related trips \\
Travel Time & Average travel time per trip (in minutes) \\
Std Travel Time & Standard deviation of trip travel times \\
First Departure Time & Hour of the first activity's departure \\
Last Departure Time & Hour of the last activity's departure \\
Home Time & Average daily duration spent at home \\
Work Time & Average daily duration spent at work \\
\bottomrule
\end{tabular}
\end{table}

\clearpage
\twocolumn
\subsection{Contextual Features}
\label{subsec: context}
The contextual attributes include: urban/rural indicator, population and housing density from the U.S. Census Bureau \cite{uscb_tiger_urban_2017, uscensus2025}, housing type and residential area proportion from the Washington State Geospatial Open Data Portal \cite{geo_wa_2025}, and both transit accessibility and network density from the EPA Smart Location Mapping dataset \cite{epa_smart_location_2025}. 

\onecolumn
\begin{table*}[!ht]
\footnotesize
\centering
\setlength{\tabcolsep}{6pt}
\caption{Contextual Features, Definitions, and Summary Statistics}
\label{tab:context}
\vspace{0.5em}
\begin{tabular}{l l l}
\toprule
\textbf{Feature} & \textbf{Definition} & \textbf{Summary Statistics} \\
\midrule
Urban Indicator & Binary variable indicating urban (1) or rural (0) & 1: 0.987,\quad 0: 0.013 \\
\midrule
Population Density & People per square mile & \begin{tabular}[t]{@{}l@{}}Mean: 16047.36 \\ Std: 14835.13\end{tabular} \\
\midrule
Distance to Transit & Distance to nearest public transit stop (m) & \begin{tabular}[t]{@{}l@{}}Mean: 270.48 \\ Std: 191.10\end{tabular} \\
\midrule
Network Density & Road length (km) per km\textsuperscript{2} of area & \begin{tabular}[t]{@{}l@{}}Mean: 29.03 \\ Std: 10.51\end{tabular} \\
\midrule
Housing Density & Housing units per square mile & \begin{tabular}[t]{@{}l@{}}Mean: 9771.24 \\ Std: 11253.57\end{tabular} \\
\midrule
Residential Proportion & Residential land share in home census block group & \begin{tabular}[t]{@{}l@{}}Mean: 0.346 \\ Std: 0.194\end{tabular} \\
\midrule
Commercial Proportion & Commercial land share in home census block group & \begin{tabular}[t]{@{}l@{}}Mean: 0.130 \\ Std: 0.114\end{tabular} \\
\midrule
Educational Proportion & Educational land share in home census block group & \begin{tabular}[t]{@{}l@{}}Mean: 0.055 \\ Std: 0.073\end{tabular} \\
\midrule
Recreational Proportion & Recreational land share in home census block group & \begin{tabular}[t]{@{}l@{}}Mean: 0.054 \\ Std: 0.094\end{tabular} \\
\midrule
Housing Type & Type of residence structure & 
\begin{tabular}[t]{@{}l@{}}
Residential Condominium: 0.235 \\
Single Family Unit: 0.334 \\
Multi-Unit (2--4): 0.059 \\
Multi-Unit ($>$5): 0.372
\end{tabular} \\
\bottomrule
\end{tabular}
\end{table*}
\clearpage

\twocolumn
\section{Additional Prediction Results}
\label{sec: income}

\subsection{Employment Status Prediction}
Similar to age prediction, employment status prediction is framed as a three-class classification task. As shown in Table~\ref{tab:employment}, while baseline models achieve reasonable overall accuracy, they struggle to accurately identify underrepresented groups (e.g., retired and employed populations) in the dataset. In contrast, our model not only achieves superior overall accuracy and weighted F1 score but also demonstrates more balanced performance across all classes.

\subsection{Income Prediction}
The prediction results for household income are presented in Table \ref{tab: income_stat}. It is important to note that the Puget Sound Regional Survey provides household-level income rather than individual income. Although we selected household representatives for analysis, their characteristics may not fully reflect household income levels due to unobserved intra-household factors such as household size. As a result, the overall prediction performance may be somewhat constrained by this limitation. Despite this limitation, our proposed model still significantly outperforms the baseline models, and demonstrates reasonable prediction performance, indicated by overall accuracy, weighted F1 score, and class-level F1 scores.

\onecolumn
\begin{table*}[ht]
\scriptsize
\centering
\caption{
Model comparison for employment status prediction. The proposed model achieves the highest overall performance and demonstrates balanced classification across categories.}
\label{tab:employment}
\begin{tabular}{lcccccc}
\toprule
Method & Class & Precision & Recall & F1-score & Total Accuracy & Weighted F1 \\ 
\midrule

\multirow{3}{*}{SVM} 
  & unemployed     & 0.167 & 0.125 & 0.143 
  & \multirow{3}{*}{0.887} & \multirow{3}{*}{0.879} \\ 
\cmidrule(lr){2-5}
  & employed & 0.930 & 0.955 & 0.943 & & \\ 
\cmidrule(lr){2-5}
  & retired   & 0.667 & 0.500 & 0.571 & & \\ 
\midrule

\multirow{3}{*}{XGBoost} 
  & unemployed     & 0.250 & 0.375 & 0.300 
  & \multirow{3}{*}{0.879} & \multirow{3}{*}{0.885} \\ 
\cmidrule(lr){2-5}
  & employed & 0.955 & 0.938 & 0.946 & & \\ 
\cmidrule(lr){2-5}
  & retired    & 0.500 & 0.250 & 0.333 & & \\ 
\midrule

\multirow{3}{*}{CatBoost} 
  & unemployed     & 0.250 & 0.250 & 0.250 
  & \multirow{3}{*}{0.887} & \multirow{3}{*}{0.886} \\ 
\cmidrule(lr){2-5}
  & employed & 0.938 & 0.946 & 0.942 & & \\ 
\cmidrule(lr){2-5}
  & retired    & 0.667 & 0.500 & 0.571 & & \\ 
\midrule

\multirow{3}{*}{GPT-4o} 
  & unemployed     & 0.276 & 1.000 & 0.432
  & \multirow{3}{*}{0.831} & \multirow{3}{*}{0.870} \\ 
\cmidrule(lr){2-5}
  & employed & 1.000 & 0.821 & 0.902 & & \\ 
\cmidrule(lr){2-5}
  & retired    & 1.000 & 0.750 & 0.857 & & \\ 
\midrule

\multirow{3}{*}{\begin{tabular}[c]{@{}c@{}}GPT-4o\\(Diaries)\end{tabular}}
  & unemployed     & 0.600 & 0.375 & 0.462 
  & \multirow{3}{*}{0.911} & \multirow{3}{*}{0.890} \\ 
\cmidrule(lr){2-5}
  & employed & 0.924 & 0.982 & 0.952 & & \\ 
\cmidrule(lr){2-5}
  & retired    & 0.000 & 0.000 & 0.000 & & \\ 
\midrule

\multirow{3}{*}{\textit{SILIC}} 
  & employed     & 1.000 & 0.375 & 0.545 
  & \multirow{3}{*}{\textbf{0.952}} & \multirow{3}{*}{\textbf{0.943}} \\ 
\cmidrule(lr){2-5}
  & unemployed & 1.000 & 0.750 & 0.857 & & \\ 
\cmidrule(lr){2-5}
  & retired    & 0.619 & 0.796 & 0.696 & & \\ 
\bottomrule
\end{tabular}
\label{tab:income}
\end{table*}

\onecolumn
\begin{table*}[ht]
\scriptsize
\centering
\caption{
Model comparison for household income prediction. The proposed model achieves the best overall and class-level performance.}
\label{tab: income_stat}
\begin{tabular}{lcccccc}
\toprule
Method & Class & Precision & Recall & F1-score & Total Accuracy & Weighted F1 \\ 
\midrule

\multirow{3}{*}{SVM} 
  & <$50k$     & 0.636 & 0.259 & 0.368 
  & \multirow{3}{*}{0.452} & \multirow{3}{*}{0.418} \\ 
\cmidrule(lr){2-5}
  & $50$--$100k$ & 0.560 & 0.255 & 0.350 & & \\ 
\cmidrule(lr){2-5}
  & $100k+$    & 0.398 & 0.833 & 0.538 & & \\ 
\midrule

\multirow{3}{*}{XGBoost} 
  & <$50k$     & 0.419 & 0.481 & 0.448 
  & \multirow{3}{*}{0.435} & \multirow{3}{*}{0.431} \\ 
\cmidrule(lr){2-5}
  & $50$--$100k$ & 0.529 & 0.327 & 0.404 & & \\ 
\cmidrule(lr){2-5}
  & $100k+$    & 0.390 & 0.548 & 0.455 & & \\ 
\midrule

\multirow{3}{*}{CatBoost} 
  & <$50k$     & 0.367 & 0.407 & 0.386 
  & \multirow{3}{*}{0.403} & \multirow{3}{*}{0.393} \\ 
\cmidrule(lr){2-5}
  & $50$--$100k$ & 0.455 & 0.273 & 0.341 & & \\ 
\cmidrule(lr){2-5}
  & $100k+$    & 0.393 & 0.571 & 0.466 & & \\ 
\midrule

\multirow{3}{*}{GPT-4o} 
  & <$50k$     & 0.333 & 0.667 & 0.444 
  & \multirow{3}{*}{0.524} & \multirow{3}{*}{0.533} \\ 
\cmidrule(lr){2-5}
  & $50$--$100k$ & 0.646 & 0.562 & 0.602 & & \\ 
\cmidrule(lr){2-5}
  & $100k+$    & 0.727 & 0.381 & 0.500 & & \\ 
\midrule

\multirow{3}{*}{\begin{tabular}[c]{@{}c@{}}GPT-4o\\(Diaries)\end{tabular}}
  & <$50k$     & 0.714 & 0.185 & 0.294 
  & \multirow{3}{*}{0.540} & \multirow{3}{*}{0.487} \\ 
\cmidrule(lr){2-5}
  & $50$--$100k$ & 0.521 & 0.909 & 0.662 & & \\ 
\cmidrule(lr){2-5}
  & $100k+$    & 0.571 & 0.286 & 0.381 & & \\ 
\midrule

\multirow{3}{*}{\textit{SILIC}} 
  & <$50k$     & 0.909 & 0.588 & 0.714 
  & \multirow{3}{*}{\textbf{0.677}} & \multirow{3}{*}{\textbf{0.678}} \\ 
\cmidrule(lr){2-5}
  & $50$--$100k$ & 0.641 & 0.610 & 0.625 & & \\ 
\cmidrule(lr){2-5}
  & $100k+$    & 0.619 & 0.796 & 0.696 & & \\ 
\bottomrule
\end{tabular}
\label{tab:income}
\end{table*}

\end{document}